# The optimality of attaching unlinked labels to unlinked meanings

Ramon Ferrer-i-Cancho[1]

ABSTRACT

Vocabulary learning by children can be characterized by many biases. When encountering a new word, children as well as adults, are biased towards assuming that it means something totally different from the words that they already know. To the best of our knowledge, the 1st mathematical proof of the optimality of this bias is presented here. First, it is shown that this bias is a particular case of the maximization of mutual information between words and meanings. Second, the optimality is proven within a more general information theoretic framework where mutual information maximization competes with other information theoretic principles. The bias is a prediction from modern information theory. The relationship between information theoretic principles and the principles of contrast and mutual exclusivity is also shown.

Keywords: vocabulary learning biases, principle of contrast, mutual exclusivity, Zipf's law, information theory

INTRODUCTION

> …what is important is the gradual development of a theory, based on a careful analysis of the… facts (...). The theory finally obtained must be mathematically rigorous and conceptually general. Its first applications are necessarily to elementary problems where the result has never been in doubt and no theory is actually required. At this early stage the application serves to corroborate the theory. The next stage develops when the theory is applied to somewhat more complicated situations in which it may already lead to a certain extent beyond the obvious and familiar. Here theory and application corroborate each other mutually. Beyond lies the field of real success: genuine prediction by theory. It is well known that all mathematized sciences have gone through these successive stages of evolution (Von Newmann & Morgenstern, 1944, pp. 7-8).

The existence of word learning biases is not in question but their origin and their fate remain unanswered (Saxton, 2010). Here new light is shed on these issues for a particular bias: when encountering a new word, children tend to assume that it means something totally different from the words that they already know (e.g., Markman & Wachtel, 1988; Merriman & Bowman, 1989; Clark, 1993). Interestingly, the same bias is found in adults (Golinkoff, Hirsh-Pasek, Bailey, & Wenger, 1992). The phenomenon is a prediction of the principle of contrast, which states that two different forms must contrast in meaning (Clark, 1987).

In essence, the problem is the following. Assume a model of semantic memory where a network defines the mapping of words into meanings. A new word arrives and the learner has two options: a) Linking that word with an unlinked meaning and b) Linking the word with a linked meaning. Strategy a) is expected from the principle of contrast (Clark, 1987) and the more restrictive principle of mutual exclusivity (e.g., Markman & Wachtel; 1988, Merriman & Bowman, 1989). Strategy a) has received the name of "disambiguation effect" (Merriman & Bowman, 1989). Despite various decades of research, the true origin of a bias for a) is still unclear (e.g., Mather & Plunkett 2009; Diesendruck & Markson, 2001; Golinkoff, Hirsh-Pasek, Bailey, & Wenger, 1992; Markman & Wachtel, 1988). Here we aim to shed light on the problem from modern information theory, show the striking convergence between information theory and theoretical perspectives from child language research concerning vocabulary learning and establish some foundations for future theoretical research.

---

[1] Complexity and Quantitative Linguistics Lab. LARCA Research Group. Departament de Ciències de la Computació, Universitat Politècnica de Catalunya (UPC). Campus Nord, Edifici Omega, Jordi Girona Salgado 1-3. 08034 Barcelona, Catalonia (Spain). Phone: +34 934134028. E-mail: rferrericancho@cs.upc.edu.

*S* is used to refer to words and *R* to refer to meanings (Ferrer-i-Cancho, 2005a; Ferrer-i-Cancho & Diaz-Guilera, 2007). *I*(*S, R*) is defined as the mutual information (Cover & Thomas, 2006) between words (*S*) and meanings (*R*) before adding the new word. Mutual information is an information theoretic measure of the capacity of words to convey meanings. Here it will be argued that the maximization of *I*(*S*,*R*) is a convenient redefinition of the principle of contrast and the principle of mutual exclusivity within the framework of information theory. $I_a'(S, R)$ and $I_b'(S, R)$ are defined as the mutual information between words and meanings after applying strategy a) and b), respectively. Here it will be shown that attaching new words to unlinked meanings is the optimal strategy with regard to mutual information because $I_a'(S, R) > I_b'(S, R)$. As there are other information theoretic principles that could be invoked, the optimality of strategy a) according to mutual information does not explain why mutual information principle is the winning principle. However, it will be shown that two other information theoretic principles in conflict with mutal information maximization, i.e. entropy minimization (Ferrer-i-Cancho & Díaz-Guilera, 2007) and compression (Chater & Vitányi, 2003; Ferrer-i-Cancho *et al*, 2013), indicate a tie between the strategies, suggesting that mutual information maximization guides vocabulary learning.

The remainder of the article is organized as follows. Section 2.1 formalizes the problem of vocabulary learning using network theory. The associations between words and meanings are regarded as links in a network. Section 2.2 presents an information theoretic model that defines mutual information (*I*(*S, R*)) assuming that all word-meaning pairs that are associated are equally likely. The next sections are based on this model. Section 2.3 analyzes the relationship between mutual information maximization and principles of language acquisition: contrast (Clark, 1987), mutual exclusivity (Markman & Wachtel, 1988) and unifunctionality (Slobin, 1985). Section 2.4 proves that attaching a new word to an unlinked meaning is optimal with respect to maximizing the mutual information of the new configuration. Section 2.5 generalizes the problem to attaching an unlinked word to $\alpha$ meanings (with $\alpha \geq 1$ and constant) and shows that the mutual information of the new configuration is maximized when the meanings are initially unlinked. Section 2.6 shows that a couple of related principles, compression (the minimization of the mean word length) and entropy minimization) are neutral in this setup in the sense that their value does not depend on the strategy. Section 2.7 generalizes the problem further with a model that combines mutual information maximization and entropy minimization. Although this model was conceived to explain the origins of Zipf's law for word frequencies, it also predicts that new cost of communication will be minimized when the new word attaches to $\alpha$ meanings that are unlinked. This is an important step in the development of a theory of communication as Von Newmann & Morgenstern (1944) indicate in the opening quote above. Finally, Section 3 discusses various aspects of the argument, from the reductions and simplifications of the real problem to a speculative connection with molecular biology.

2. INFORMATION THEORY OF VOCABULARY LEARNING

2.1 The lexical matrix

The mathematical framework is abstract enough to allow one to replace words by forms, labels, tags, gestures, signals, … and so on and to replace meanings by concepts, synsets, objects, extensions, categories and… so on. Assume that words of the potential vocabulary and meanings of the potential repertoire of meanings are indexed with integer numbers from 1 to $V_S^{max}$ and numbers from 1 to $V_R^{max}$, respectively. The new word learning scenario can be formalized by means of a lexical matrix $A=\{a_{ij}\}$ (Hurford, 1989; Steels, 1995; Nowak & Komarova, 2001), where $a_{ij} = 1$ if the *i*-th word is connected with the *j*-th meaning ($a_{ij} = 0$ otherwise). This matrix indeed defines the connections of a bipartite network which can be seen as a model of semantic memory (Baronchelli, Ferrer-i-Cancho, Pastor-Satorras, Chater, & Christiansen, 2013). $\mu_i$ is defined as the degree of the *i*-th word (the degree of a word or a meaning is its number of connections (Baronchelli, Ferrer-i-Cancho, Pastor-Satorras, Chater, & Christiansen, 2013) and $\omega_i$ is the degree of the *i*-th meaning, thus

$$\mu_i = \sum_{j=1}^{V_R^{max}} a_{ij} \text{ and } \omega_i = \sum_{j=1}^{V_S^{max}} a_{ji} \tag{1}$$

and the number of connections is

$$M = \sum_{i=1}^{V_S^{max}} \mu_i = \sum_{i=1}^{V_R^{max}} \omega_i. \tag{2}$$

For simplicity, our framework does not take into account that meanings are structured. For instance, meanings (concepts) are interrelated and believed to be organized hierarchically (e.g., Fellbaum & Miller (1998)). However, our model is not incompatible with a taxonomic structure. The concept associated to the word "animal" might be implemented in our framework as many connections from the word "animal" to meanings corresponding to low level concepts of mammal such as "dog", "horse", and so on. Our set of meanings may also include categories and their generalizations ("dog","horse" and "animal" at the same time). Certain meanings may differ only in taxonomic depth but our level of abstraction allows for a totally flat taxonomic structure (all meanings being at the same taxonomic level).

Assume that there are $V_S$ words with at least one connection and $V_R$ meanings with at least one connection. Without any loss of generality, one can assume that the first $V_S$ words are the linked words, i.e.

$$\mu_1,...,\mu_i,...,\mu_{V_S} > 0 \text{ and } \mu_{V_S+1}, \mu_{V_S+2},... = 0 \tag{3}$$

and that the first $V_R$ meanings are the linked meanings, i.e.

$$\omega_1,...,\omega_i,...,\omega_{V_R} > 0 \text{ and } \omega_{V_R+1},\omega_{V_R+2},... = 0. \tag{4}$$

A new word is an unlinked word. Hereafter, it is assumed that learning a new word is possible, i.e. $V_S < V_S^{max}$. Without any loss of generality, let us assume that the new word that arrives is the ($V_S$+1)-th. Assuming that $V_S > 0$, the learner has to link the ($V_S$+1)-th word with a meaning of degree $k_R$, that changes its degree to $k_R$+1. Let $j$ be the index of the candidate meaning. The matrix $A$ is updated changing $a_{V_S+1,j} = 0$ to $a_{V_S+1,j} = 1$. The learner has two options for updating $A$:

a) $k_R = 0$ (principle of contrast), i.e., linking the ($V_S$+1)-th word with an unlinked meaning. Without any loss of generality, let us assume that the meaning is the ($V_R$+1)-th.
b) $k_R > 0$, i.e. linking the ($V_S$+1)-th word with a linked meaning.

Our approach is similar to that of statistical mechanics in physics: we do not aim to provide a detailed model of vocabulary learning but a simplified model of reality that captures, qualitatively a wide range of phenomena, which not only comprises word learning biases (the original contribution of this article) but also statistical properties of word frequencies such as Zipf's law (Ferrer-i-Cancho, 2005a). Indeed, the model is reminiscent of the Ising model that has been used successfully in statistical mechanics as a simplified model for phase transitions between ferromagnetic and paramagnetic states (Pang, 2006; Kobe, 2000). As the Ising model reduces the magnetic moments to only two positions "up" and "down", our lexical matrix approach reduces the weight of a semantic connection between a word and a meaning to zero and one. As the dynamics of the system in the Ising model is guided by the minimization of a Hamiltonian function (Pang, 2006), it will be argued the learning of new words by children and adults is guided to a large extent by the maximization of a mutual information (or a generalization including other principles, e.g., Ferrer-i-Cancho & Díaz-Guilera 2007) defined on a lexical matrix of binary states. This kind of minimalistic approach is also found in successful research on the crucial role of the topology of social interactions for the dynamics of collective naming games (Baronchelli, Dall'Asta, Barrat, & Loreto, 2007).

2.2. Information theory of the lexical matrix.

The mutual information between words and meanings, $I(S,R)$, can defined in a convenient way as

$$I(S,R) = H(S) - H(S|R). \tag{5}$$

$H(S)$ is the entropy of words and $H(S|R)$ is the conditional entropy of words given meanings (Ferrer-i-Cancho & Díaz-Guilera, 2007).

The idea that word frequency is an epiphenomenon of the mapping of words into meanings has been used to shed light on the origins of Zipf's law for word frequencies (Manin, 2008). A family of information theoretic models of that law (Ferrer-i-Cancho, 2005a; Ferrer-i-Cancho, 2005b), assumes that $M$ is finite (e.g., through finite $V_S^{max}$ and $V_R^{max}$) with $M>0$ and that the joint probability of the $i$-th word and the $j$-th meaning is

$$p_{ij} = \frac{a_{ij}}{M}. \tag{6}$$

Eq. 6 is the fundamental assumption of a family of simple models of Zipf's law for word frequencies (Ferrer-i-Cancho, 2005a; Ferrer-i-Cancho, 2005b) which yields

$$H(S) = \log M - \frac{1}{M} \sum_{i=1}^{V_S^{max}} \mu_i \log \mu_i \tag{7}$$

and

$$H(S|R) = \frac{1}{M} \sum_{i=1}^{V_R^{max}} \omega_i \log \omega_i. \tag{8}$$

Thanks to the convention 0log0 = 0 (Cover & Thomas, 2006) and the arrangement of words and meanings described in Eqs. 3-4, Eqs. 7-8 become

$$H(S) = \log M - \frac{1}{M} \sum_{i=1}^{V_S} \mu_i \log \mu_i \tag{9}$$

and

$$H(S|R) = \frac{1}{M} \sum_{i=1}^{V_R} \omega_i \log \omega_i. \tag{10}$$

Applying Eqs. 9-10 to Eq. 5, it is obtained

$$I(S,R) = \log M - \frac{1}{M} \left( \sum_{i=1}^{V_S} \mu_i \log \mu_i + \sum_{i=1}^{V_R} \omega_i \log \omega_i \right). \tag{11}$$

2.3. The relationship between mutual information maximization and principles from the field of language acquisition.

The principle of contrast states that "*every two forms contrast in meaning*" (Clark, 1987) while the principle of mutual exclusivity states that "*each object could have only one category label, and each label could refer to only one category of objects*" (Markman & Wachtel, 1988; see also Merriman & Bowman, 1989). Notice that this definition mixes "referents" (objects) and "categories" while contrast stays at the level of categories. An object can be labeled with a word but this relationship is mediated by a concept or category, at least in human language. To ease a comparison among different principles, we focus on a less restrictive version of mutual exclusivity that is defined on categories, i.e. "*each category of objects could have only one label, and each label could refer to only one category of objects*". This redefined principle of mutual exclusivity is clearly similar to the principle of unifunctionality, according to which one-to-one mappings of forms into meanings in children are favoured (Slobin, 1985). In a strict sense, one-to-one mappings are particular cases of mutual exclusivity where no word or meaning is left unlinked. In order to allow for a fairer comparison with the principle of contrast, we are also generalizing the original principle of mutual exclusivity allowing for categories of different taxonomic levels.

$I(S,R)$ is an information theoretic measure of the capacity of words to convey meanings. Intuitively, the principle of mutual information maximization promotes that words become identifiers of meanings. When $V_S^{max}=V_R^{max}$, mutual information maximization predicts a one-to-one mapping of words into meanings, that is, words are true identifiers of meanings (Ferrer-i-Cancho & Díaz-Guilera, 2007; Appendix A). In this context, the principle is equivalent to the principle of contrast and the principle of mutual exclusivity if disconnected vertices are not allowed. However, when $V_S^{max}<V_R^{max}$, it is no longer necessary that a word connects with only one meaning in order to maximize $I(S,R)$ (Ferrer-i-Cancho & Díaz-Guilera, 2007; Appendix A). For instance, let us assume that $V_S^{max} = V_S = 2$ and $V_R^{max} = 4$ and consider two different configurations: one where every word connects with a single meaning exclusively (Fig. 1 (a)) and another where each word is connected with a couple of meanings exclusively (Fig. 1 (b)). Here "exclusively" means that the meanings of words do not overlap (a meaning cannot be linked to more than one word). Notice that the configuration in Fig. 1 a) satisfies both the definition of the principle of contrast and that of mutual exclusivity in a strict sense but that of Fig. 1 b) only satisfies the principle of contrast. Interestingly, the mappings of Figs. 1 a) and b) maximize mutual information[2]. Thus, the principle mutual exclusivity in a narrow sense is not compatible with optimal solutions according to the general principle of mutual information maximization. When $V_S^{max} > V_R^{max}$, it is not necessary that a meaning connects to just one word in order to maximize the mutual information[3]. Thus, neither the principle of contrast nor that of mutual exclusivity are compatible with mappings maximizing mutual information when $V_S^{max} > V_R^{max}$. Interestingly, contrast in Clark's sense (Clark, 1987) is more compatible with mutual information maximization than mutual exclusivity.

A pitfall of the principle of contrast and mutual exclusivity is that they do not warrant that enough information is transmitted. Consider again that $V_S^{max} = 2$ and $V_R^{max} = 4$ with $M = 1$ (Fig. 1 (c)), then Eq. 11 gives $I(S,R)=0$ because all the degrees are zero. The information transmitted is minimum as $I(S,R) = 0$ but the definition of the principle contrast is satisfied (every meaning has at most one form) and the same applies to that of mutual exclusivity (every form and every meaning have at most one connection). The principle of contrast is a necessary but not sufficient condition for maximum $I(S,R)$ only when $V_S^{max} \leq V_R^{max}$ (Appendix A). The principle of mutual exclusivity is only needed by maximum $I(S,R)$ when $V_S^{max} = V_R^{max}$ (Appendix A).

Mutual information maximization can be seen as reformulation of the principle of contrast or that of mutual exclusivity that is naturally supported by information theory. While the principle of contrast and that of mutual exclusivity need another principle to warrant the mapping will informative, mutual information maximization does it in one shot and thus offers a more parsimonious explanation.

---

[2] In the first configuration (Fig. 1 (a)), one has $V_R = 2$, $M = 2$, $\mu_1 = \mu_2 = 1$, and $\omega_i \in \{0,1\}$ for $i = 1,...,4$, which according to Eqs. 9 and 11 and the 0log0 = 0 convention gives $I(S,R) = H(S) = \log 2$. In the second configuration (Fig. 1 (b)), one has $V_R = 4$, $M = 4$, $\mu_1 = \mu_2 = 2$, $\omega_i = 1$ for $i = 1,...,4$, which gives $I(S,R) = H(S) = \log 2$ again. $I(S,R)$ is maximum in both cases because $I(S,R) = H(S)$ and $H(S)$ is maximum because $H(S) \leq \log V_S^{max} = \log 2$, the maximum value that $H(S)$ can reach.

[3] This is easy to see by inverting the color of circles in Fig. 1. b) and computing $I(S,R)$ again. For a general explanation, see Ferrer-i-Cancho & Díaz-Guilera (2007) and Appendix A.

2.4. The maximization of I(S,R) predicts that new words are attached to unlinked meanings.

$V'_S$, $V'_R$, $\mu'_i$, $\omega'_i$ and $M'$ are used to refer to values of $V_S$, $V_R$, $\mu_i$, $\omega_i$ and $M$ after updating the matrix $A$ following strategy a) or b). $I_x'(S, R)$ and $H_x'(S)$ are defined as the mutual information between words and meanings and the word entropy after updating the matrix $A$ following strategy $x$. Eq. 9 gives

$$H'_a(S) = H'_b(S) = \log(M') - \frac{1}{M'}\sum_{i=1}^{V'_S} \mu'_i \log \mu'_i. \tag{12}$$

Applying $V'_S = V_S+1$, $M' = M+1$, $\mu'_i = \mu_i$ for $1 \leq i \leq V_S$ and $\mu'_{V_S+1} = 1$, Eq. 12 becomes

$$H'_a(S) = H'_b(S) = \log(M+1) - \frac{1}{M+1}\sum_{i=1}^{V_S} \mu_i \log \mu_i. \tag{13}$$

Let $\alpha_0$ be a boolean parameter indicating if the target meaning is unlinked. Then one has that $V'_R = V_R + \alpha_0$, $\omega'_{V_R+1} = \alpha_0$ and then Eq. 10 gives

$$H'(S \mid R) = \tag{14}$$

$$\frac{1}{M'}\sum_{i=1}^{V_R+\alpha_0} \omega'_i \log \omega'_i = \frac{1}{M+1}\left(\sum_{i=1}^{V_R} \omega'_i \log \omega'_i + \alpha_0 \log \alpha_0\right) = \frac{1}{M+1}\sum_{i=1}^{V_R} \omega'_i \log \omega'_i$$

For strategy b), Eq. 14 yields

$$H_b'(S \mid R) = \frac{1}{M+1}\left(\sum_{i=1}^{V_R} \omega_i \log \omega_i + \delta(k_R)\right) \tag{15}$$

with

$$\delta(k) = (k+1)\log(k+1) - k \log k. \tag{16}$$

Recall that $k_R$ is the initial degree of the target meaning. For strategy a), $\omega'_i = \omega_i$ for $1 \leq i \leq V_R$ transforms Eq. 14 into

$$H_a'(S \mid R) = \frac{1}{M+1}\sum_{i=1}^{V_R} \omega_i \log \omega_i = \frac{1}{M+1}\left(\sum_{i=1}^{V_R} \omega_i \log \omega_i + \delta(0)\right) \tag{17}$$

as $\delta(0)=0$ thanks to the 0log0=0 convention. Thus, $H_a'(S \mid R)$ is a particular case of $H_b'(S \mid R)$ with $k_R = 0$. Combining Eqs. 15 and 17 it is obtained

$$H_b'(S \mid R) = H_a'(S \mid R) + \Delta_{H(S \mid R)} \tag{18}$$

with

$$\Delta_{H(S|R)} = \frac{\delta(k_R)}{M+1}. \tag{19}$$

As $M > 0$ and $\delta(k_R) > 0$ thanks to $k_R > 0$ for that strategy, one has $\Delta_{H(S|R)} > 0$ and thus $H_a'(S|R) < H_b'(S|R)$. The definition of $I(S,R)$ in Eq. 5, $H_a'(S) = H_b'(S)$ (recall Eq. 13) and Eq. 18 imply

$$I_a'(S,R) = I_b'(S,R) + \Delta_{H(S|R)}. \tag{20}$$

Thus, $I_a'(S|R) > I_b'(S|R)$ as $\Delta_{H(S|R)} > 0$. This proves that attaching a new word to unlinked meanings is optimal from the perspective of mutual information maximization.

2.5. A generalized scenario with multiple connections.

In the scenario of the preceding subsection, only one connection is added (in both strategies). This means that $M' = M + 1$ and words cannot be polysemous or have broad or multiple extensions. A generalization consists of α connections to "meanings" from the new word.

The case of $M' = M + α$, with $α > 0$ is considered next. Under a narrow interpretation of the principle of mutual exclusivity, one may think that the possibility that $α > 1$ contradicts the assumption that *"each label could refer to only one category of objects"* (Markman & Wachtel, 1988) but here mutual information maximization is regarded as the fundamental force, which is not totally incompatible with mutual exclusivity as it is explained in Section 2.3. Interestingly, $α > 1$ is not a problem for the principle of contrast.

It will be shown that linking with disconnected meanings when $α > 0$, a generalized strategy a), is still the optimal strategy. Let us consider a generalized scenario where the new word has to be attached to an arbitrary set of α meanings (no constraint is imposed on whether those meanings are linked or unlinked). It is known that the degrees of the meanings within that set are $k_1, k_2, ..., k_α$ ($k_1, k_2, ..., k_α \geq 0$) and that $α_0$ of those meanings are unlinked, with $0 \leq α_0 \leq α$.

The word entropy after updating $A$ is

$$H'(S) = \log(M+\alpha) - \frac{1}{M+\alpha}\left(\sum_{i=1}^{V_S+1} \mu'_i \log \mu'_i\right) \tag{21}$$
$$= \log(M+\alpha) - \frac{1}{M+\alpha}\left(\sum_{i=1}^{V_S} \mu_i \log \mu_i + \alpha \log \alpha\right).$$

Notice that $H'(S)$ does not depend on $k_1, k_2, ..., k_α$ and thus does not depend on the strategy chosen to form the $α$ new links as it occurred for the particular case of $α = 1$ in Section 2.4.

As $V'_R = V_R + α_0$, the word conditional entropy after updating $A$ is

$$H'(S|R) = \frac{1}{M+\alpha}\left(\sum_{i=1}^{V_R+\alpha_0} \omega'_i \log \omega'_i\right) = \frac{1}{M+\alpha}\left(\sum_{i=1}^{V_R} \omega_i \log \omega_i\right) + \Delta_{H(S|R)}, \tag{22}$$

being

$$\Delta_{H(S|R)} = \frac{1}{M+\alpha} \sum_{i=1}^{\alpha} \delta(k_i) \tag{23}$$

the contribution of the particular strategy adopted to $H'(S|R)$.

Keeping $\alpha$ constant, $\Delta_{H(S|R)}$ (and thus $H'(S|R)$) is minimized by $k_1, k_2, ..., k_\alpha = 0$ (notice that $\delta(k_i) \geq \delta(0)$ as $k_i \geq 0$). This means that $H'(S|R)$ is minimized when $\alpha_0 = \alpha$, i.e. all the new links are formed with disconnected meanings. As $I'(S,R) = H'(S) - H'(S|R)$ and $k_1, k_2, ..., k_\alpha$ only have an influence on $H'(S|R)$, we conclude that $I'(S,R)$ is maximized when $k_1, k_2, ..., k_\alpha = 0$. This indicates that attaching an unlinked word only to unlinked meanings is optimal in terms of mutual information maximization.

2.6. The neutrality of other competing information theoretic principles.

It has been shown above that the principle of mutual information maximization could explain why new words are attached to unlinked meanings if it is the only principle at work. However, other principles might be competing or be in conflict with mutual information maximization. If that was the case, one should explain why mutual information maximization is still the winner.

An important information theoretic principle is compression (Chater & Vitányi, 2003), which is formalized as the minimization of the mean code length (Ferrer-i-Cancho *et al*, 2013). The mean coding length of a repertoire *S* is defined as

$$L(S) = \sum_{i=1}^{V_S} p(s_i) l(s_i), \tag{24}$$

where $p(s_i)$ is the probability of the *i*-th signal and $l(s_i)$ is its length in bits. The law of brevity, the tendency of more frequent words has been argued to be an epiphenomenon of the minimization of $L(S)$ (Ferrer-i-Cancho *et al*, 2013). For that reason, the principle of minimization of $H(S)$ can be seen as a particular principle of compression under optimal uniquely decipherable coding and variable probabilities. More precisely, Ferrer-i-Cancho *et al*, 2013 consider a generalization of $L(S)$ where lengths cannot only be measured in bits, nats but also letters or time. That generalized $L(S)$ is assumed to be proportional to $L(S)$.

Another important principle is word entropy minimization (Ferrer-i-Cancho, 2005a; Ferrer-i-Cancho & Díaz-Guilera, 2007). A critical balance between the maximization of $I(S,R)$ and the minimization of $H(S)$ is believed to underlie Zipf's law for word frequencies (Ferrer-i-Cancho & Solé, 2003; Ferrer-i-Cancho, 2005a). The conflict between mutual information maximization and the minimization of word entropy or mean code length is easy to see from the following chain of inequalities (Cover & Thomas, 2006)

$$I(S,R) = H(S) - H(S|R) \leq H(S) \leq L(S). \tag{25}$$

The inequality $H(S) \leq L(S)$ holds for uniquely decipherable codes, which covers a wide class of coding schemes (Cover & Thomas, 2006). The fact that $H(S)$, $H(S|R)$ and $I(S, R)$ are non-negative means that $I(S,R) \leq H(S)$ and thus the maximization of $I(S, R)$ and the minimization of $H(S)$ are forces in conflict. The same conflict applies to $L(S)$ for being an upper bound of $H(S)$ and thus an upper bound of $I(S, R)$.

After updating the matrix, both strategies give the same mean code length (Cover & Thomas, 2006), i.e.

$$L'_a(S) = L'_b(S) = \sum_{i=1}^{V_S+1} p'(s_i) l(s_i), \qquad (26)$$

where $p'(s_i)$ is the new probability of $s_i$ after applying one of the strategies. In our lexical matrix framework (Ferrer-i-Cancho & Díaz-Guilera 2007),

$$p(s_i) = \frac{\mu'_i}{M'}, \qquad (27)$$

and thus $p(s_i) = \mu_i/(M + \alpha)$ if $1 \leq i \leq V_S$ and $p(s_i) = \alpha/(M + \alpha)$ if $i = V_S+1$.

Eq. 26 indicates that the tie between both strategies concerning the minimization of $L(S)$ illustrated by Eq. 26 is broken by the strategy that maximizes mutual information, namely, that of attaching new labels to unlinked meanings. The same tie applies to word entropy minimization as it has been shown above that $H'_a(S) = H'_b(S)$ (Sections 2.4 and 2.5). In sum, both compression and entropy minimization are in conflict with mutual information maximization but are neutral concerning the best linking strategy.

2.7. Attaching new words to unlinked meanings is predicted by a model of Zipf's law.

Our findings on the optimality of strategy a) show the predictive power of a model of word frequencies that was not originally conceived to explain vocabulary learning biases. This model addresses the problem of how a communication would integrate the principle of mutual information maximization and the principle of word entropy minimization by assuming that both are combined linearly to give the cost function (Ferrer-i-Cancho & Díaz-Guilera, 2007)

$$\Omega(\lambda) = -\lambda I(S, R) + (1 - \lambda) H(S), \qquad (28)$$

where $\lambda$ is a parameter controlling for the weight of each of the two pressures ($0 \leq \lambda \leq 1$). If $\lambda=0$ then mutual information maximization is irrelevant and if $\lambda=1$ then word entropy minimization is irrelevant.

Plugging $I(S,R)=H(S) - H(S|R)$ into Eq. 28, it is obtained (Ferrer-i-Cancho & Díaz-Guilera, 2007)

$$\Omega(\lambda) = (1 - 2\lambda) H(S) + \lambda H(S | R). \qquad (29)$$

Now, let us consider that a learner is faced with the problem of having to assign meanings to a new word. Imagine that the lexical matrix is updated so that the new value of $\Omega(\lambda)$, i.e.

$$\Omega'(\lambda) = (1 - 2\lambda) H'(S) + \lambda H'(S | R), \qquad (30)$$

is as small as possible. The value of $\lambda$ is *a priori* unknown. However, for the arguments that follow its value is irrelevant provided that the maximization of *I(S,R)* is not inhibited *a priori*, which is warranted for $\lambda>0$. Thus, let us assume that $\lambda>0$.

As *H'*(*S*) does not depend on the strategy used, the best $\Omega'(\lambda)$ is determined by the strategy giving the smallest value of *H'*(*S|R*), which is a). Thus, strategy a) is also optimal in terms of the minimization of $\Omega'(\lambda)$. Interestingly, the conclusion is robust in terms of how *H*(*S*) minimization and *I*(*S,R*) maximization are combined. Thus, a simple information theoretic model is able to shed light on two apparently unrelated phenomena: Zipf's law for word frequencies (Ferrer-i-Cancho, 2005a) and choices during vocabulary learning.

3. DISCUSSION

Our theoretical framework is able to deal with the taxonomic organization of word meanings, from concrete concepts such as dog or horse to generalizations such as mammal or animal. Imagine that the repertoire of meanings includes concepts of varying taxonomic depth as those mentioned above. The content of this repertoire of meanings assumes that the child has realized that meanings can also contrast in taxonomic depth, something that does not happen initially (Clark, 1987 and references therein). Imagine that a child is presented with the label "animal" when he/she has not actually learned its meaning but he/she has a word for dog, a word for horse, and so on. This is a plausible situation due to the bias for low taxonomic levels in vocabulary learning (Mervis, 1987). According to that bias, the words for horse and dog are likely to have been learned before the words for animal because they are at a lower level in the hierarchy of concepts. The principle of mutual information maximization favours linking the word "animal" to a concept that does not cover exactly a dog, a horse,…This suggests that mutual information maximization could facilitate the learning of abstract words when combined with the basic level bias. However, the extension of "animal" includes that of dog, horse,…and that could turn the learning of the meaning animal harder. Future research should address with more detail the problem of learning of words with meanings that are taxonomically related with the meaning of already learned words.

If strategy a) with $\alpha=1$ was the only mechanism by which the lexical matrix *A* is updated, words would never be polysemous. There are various ways of producing polysemous words. One is obviously $\alpha>1$. Another one are updates involving currently linked words ("old" words). Our point here is that changes involving "old" words might be necessary to minimize *H*(*S*). Imagine that $\alpha$ new links are added to an "old" word that has degree $k_S$ (for the word to be "old", $k_S > 0$). We generalize the definition of $\delta(k)$ in Eq. 16 as

$$\delta(k, \alpha) = (k + \alpha) \log (k + \alpha) - k \log k. \qquad (31)$$

After updating the matrix, the new entropy becomes

$$H''(S) = \log(M + \alpha) - \frac{1}{M + \alpha}\left(\sum_{i=1}^{V_S} \mu_i \log \mu_i + \delta(k_S, \alpha)\right). \qquad (32)$$

and then (recall Eq. 21)

$$H'(S) = H''(S) + \Delta_{H(S)} \tag{33}$$

with

$$\Delta_{H(S)} = \frac{1}{M+\alpha}(\alpha \log \alpha + \delta(k_S, \alpha)). \tag{34}$$

It is easy to see that $\Delta_{H(S)} > 0$ (as $\alpha$, $k_S > 0$) and thus $H'(S) > H''(S)$. We conclude that increasing polysemy is expected by the principle of entropy minimization.

It has not been specified so far if the words are connected to linked or unlinked meanings. Let us consider the particular case of linking to new meanings. The fact that $H'(S) > H''(S)$ implies that a communication system might prefer to reach the unlinked words recycling old words. However, notice that a system covering unlinked meanings only by adding a new word with α links to maximize $I(S,R)$ is paying the maximum cost for communicating according to $H(S)$ for a given $V_S$ as all the linked words are equally likely, which gives $H(S) = \log V_S$ (Ferrer-i-Cancho & Díaz-Guilera, 2007). That might be affordable initially, namely when $V_S$ is small, but might not be sustainable in the long run. At some point, a struggle between maximization of $I(S,R)$ by adding new words and recycling old words to minimize $H(S)$ is expected. Thus, the addition of new words cannot be the only track to reach unlinked meanings.

The minimization of $H(S)$ (recall $H'(S) > H''(S)$) puts pressure towards recycling "old" words to reach not only unlinked but also linked meanings. This pressure to recycle might explain why children overextend word meanings so frequently. For children of about two years, overextensions account for about 40% of words uses (Rescorla, 1980).

Research indicates that 'mutual exclusivity' is overridden by children at some point (Markman, 1990; Imai & Haryu, 2004) but it does not disappear completely (Davidson & Tell, 2005). According to information theory, the traditional definition of the bias might be overridden at least by two forces: pressure to minimize $H(S)$ (recall that $H'(S) > H''(S)$) but also pressure to maximize $I(S,R)$. As for the latter, recall that the optimal configurations are only one-to-one mappings if the size of the repertoire of words and that of meanings is the same (Appendix A). If the repertoire of meanings is larger, words can be attached to multiple meanings and mutual information still be optimum (recall Fig. 1 (b) and Section 2.3). This situation might apply to children at some point as they might not be able learn words with the same speed as their repertoire of meanings grows. Although mutual exclusivity has to be over-ridden sometime, the principle of contrast can accommodate pressure for minimizing $H(S)$ and maximizing $I(S,R)$ when the number of meanings exceeds by far the number available words. Not surprisingly then, contrast remains: it applies to children and adults alike (Clark, 1987). This might have an information theoretic explanation: the principle of contrast is needed both by maximum $I(S,R)$ and minimum $H(S)$ (Appendix A and B).

Mutual information maximization might provide a more parsimonious account of vocabulary learning biases. The 'mutual exclusivity' bias is problematic for penalizing configurations with maximum mutual information transfer and for being overridden in more circumstances than with an approach based on mutual information maximization. The latter approach has the additional virtue of allowing one to recur coherently to the minimization of $H(S)$, a closely related information theoretic metric, for covering what mutual information maximization cannot explain, thus connecting with a powerful information theory framework (Ferrer-i-Cancho & Diaz-Guilera, 2007).

In her pioneering research Clark, proposed that the principle of contrast should lie in children's earlier recognition of adult intentions as a part of rational behavior (Clark, 1988, pp. 324). In order to select strategy a), the child should make a pragmatic inference about the speaker's intention: the speaker would have used an old word if he was referring to and object of a category for which a label already exists. The crux of that rational behavior could reduce simply to the maximization of the mutual information between words and something else (e.g., concepts, categories, objects,…). Such "rational behavior" could be an epiphenomenon of a rather simple optimization principle rather than the result of a complex cognitive process within an individual. If such rational behavior reduces to maximizing that mutual information, it may not be so important that children realize about adult intentions or adult pragmatic directions (Clark & Grossman, 1998). A question for future research is if mutual information maximization or contrast and related principles are possible without rationality or a mind. A challenging example of that is the mapping of amino acids into codons in DNA sequences. Every two aminoacids map to different codons thus, it can be said that this mapping obeys a generalized principle of contrast in the sense that "two amino acids contrast in their codon". Furthermore, the relationship between codons and amino acids resembles the mapping of words into meanings in the sense that it looks arbitrary (Bel-Enguix & Jiménez-López, 2011). Thus one could envisage a process where a new amino acid arrives and one has to decide that a) the new amino acid is attached to an unlinked codon or that b) the new amino acid is attached to a linked codon. It looks as if strategy a) had been selected as no codon leads to more than one amino acid. A challenge for this view is not only the plausibility of the process but the fact that it leads to some equivalence between words and amino acids while it has been argued the opposite, namely that words are equivalent to codons (Bel-Enguix & Jiménez-López, 2011). This should be investigated further in the future.

It has been shown that attaching new words to unlinked meanings is predicted by the principle of mutual information maximization for a simple model based on a lexical matrix. This provides further support for the hypothesis that mutual information maximization is a fundamental principle of communication, with human language as a particular case. That principle has been combined with a principle of word entropy minimization to explain the origins of Zipf's law for word frequencies in languages (Ferrer-i-Cancho 2005a; Ferrer-i-Cancho & Solé, 2003). Thus, mutual information maximization might be "*the powerful memory mechanism, enabling mutual exclusivity to emerge across repeated exposures to potential referents*" (Mather & Plunkett, 2009). Only that abstract information theoretic principle might explain the robustness of the avoidance of synonymy (Clark, 1987; Manin, 2008), which concerns both children and adults (Golinkoff, Hirsh-Pasek, Bailey, & Wenger, 1992), never disappears entirely (Davidson & Tell, 2005), surfaces across languages (Imai & Haryu, 2004), manifests even when referents are not visible (Markman *et al*, 2003) and goes beyond naming (Diesendruck & Markson, 2001). The apparent universality of the avoidance of synonymy (Imai & Haryu, 2004) might be simply a consequence of the fact that all brains, regardless of the languages they host, solve the associative learning problem of mapping words into 'meanings'. Mutual information is just measuring the quality of this learning. Associative learning is the hypothesis to favour over the less parsimonious hypothesis that the bias has primarily a genetic origin, in agreement with Imai & Haryu (2004). Universal properties of the brain or cognition are the minimalist key to understanding universal linguistic phenomena (Ferrer-i-Cancho, 2006; Ferrer-i-Cancho, 2005a; Ferrer-i-Cancho *et al*, 2013).

In the present article, it has just been proven the optimality of attaching new labels to unlinked meanings within a specific framework. There is still a rich phenomenology awaiting to be illuminated by information theory: e.g., the rise of the bias by the age of 18-22 months (Halberda, 2003; Mather & Plunkett, 2009) and its fate (e.g., Imai & Haryu, 2004; Davidson & Tell, 2005), the interaction with other vocabulary learning biases (e.g., Ima & Haryu, 2004), or the even more detailed evolution that can be observed in ultradense corpora of the visual and auditory experience of a child (Roy *et al*, 2006; Roy 2009). This article has established some mathematical foundations for future theoretical research. Beyond child language, our framework could help to understand the evolution of word meaning (Manin 2008).

ACKNOWLEDGEMENTS

We are grateful to B. Elvevåg, M. Christiansen, E. Clark, Ł. Debowski, A. Hernández-Fernández and G. Wimmer for helpful comments. This research was supported by the grant BASMATI (TIN2011-27479-C04-03) from the Spanish Ministry of Science and Innovation, the grant APCOM (TIN2014-57226-P) from MINECO (Ministerio de Economía y Competitividad) and the grant 2014SGR 890 (MACDA) from AGAUR (Generalitat de Catalunya).

APPENDIX

A. The mappings maximizing mutual information.

Here the optimal mappings of words into meanings are studied in two cases: $V_S^{max} \leq V_R^{max}$ and $V_S^{max} \geq V_R^{max}$. $V_S^{max} \leq V_R^{max}$ looks appropriate for human language. $V_S^{max} \geq V_R^{max}$ looks appropriate for the correspondence between codons and amino acids under the hypothesis that codons are playing the role of words (Bel-Enguix & Jiménez-López, 2011), $V_S^{max} = 4^3 = 64$ and $V_R^{max} = V_R = 20$ ($V_S^{max} - V_S = 3$ due to the three stop codons that do not code for any amino acid).

Suppose a mapping of words into meanings that satisfies the following conditions:

1. $\mu_i = k$ for $i = 1,2,...V_S^{max}$ for some natural number $k \geq 1$.
2. $\omega_i \in \{0,1\}$ (principle of contrast).

Notice that the first condition imposes $V_S = V_S^{max}$ while the second condition imposes $kV_S^{max} \leq V_R^{max}$ and thus

$$1 \leq k \leq \lfloor V_R^{max}/V_S^{max} \rfloor. \tag{A.1}$$

Notice also that $k > 0$ is needed by $M>0$. The existence of those maximal mappings needs that $V_S^{max} \leq V_R^{max}$. Otherwise no $k$ satisfies Eq. A.1.

We will show that those mappings define the configurations maximixing $I(S,R)$ when $V_S^{max} \leq V_R^{max}$ in two steps. First, showing that those configurations yield maximum $I(S,R)$. Second, showing that they are the only configurations.

First step: notice that the second condition implies $H(S|R) = 0$ according to Eq. 8 thanks to the 0log0 = 0 convention. Applying the first condition and $M = kV_S^{max}$, it is obtained $H(S) = \log V_S^{max}$ (form probability is proportional to degree and thus all forms have probability $1/V_S^{max}$). Thus, $H(S)$ is taking its maximum possible value whereas $H(S|R)$ is taking its minimum value. As $I(S,R) = H(S)-H(S|R)$, it follows that $I(S, R)$ is maximum. Eq. A.1 implies that one-to-one mappings are the only mappings maximizing $I(S, R)$ when $V_S^{max} = V_R^{max}$.

Second step: notice that

- If condition 1 fails, then words are not equally likely as the probability of a word is proportional to its degree. Then one has that $H(S) < \log V_S^{max}$ and it follows that $I(S,R)$ is not maximum because $I(S,R) \leq H(S)$.
- If condition 2 fails, then $H(S|R) > 0$ and thus $I(S,R) < \log V_S^{max}$ even if $H(S)$ is maximum because $I(S,R) = H(S)-H(S|R)$.

By symmetry, the mappings of words into meanings that satisfy the following conditions:

1. $\omega_i = k$ for $i = 1,2,...V_R^{max}$ for some natural number $k \geq 1$.
2. $\mu_i \in \{0,1\}$.

are the configurations that maximize $I(S,R)$ when $V_S^{max} \geq V_R^{max}$.. As before, first condition imposes $V_R = V_R^{max}$ while the second condition implies

$$1 \leq k \leq \lfloor V_S^{max}/V_R^{max} \rfloor. \tag{A.2}$$

The latter is why the existence of these maxima needs that $V_S^{max} \geq V_R^{max}$.

In sum, the principle of contrast is a necessary condition for maximum $I(S,R)$ only when $V_S^{max} \leq V_R^{max}$. The principle of contrast is not a sufficient condition for maximum $I(S,R)$ in that case. A mapping where words do not have the same degree, will give a value of $I(S,R)$ that is not maximum as $I(S,R)$ is bounded above by $H(S)$ and maximum $H(S)$ needs that all the form degrees are the same. Fig. 1 (c) shows a mapping satisfying the principle contrast with minimum $I(S,R)$. The principle of mutual exclusivity is more demanding. It is only needed by maximum $I(S,R)$ when $V_S^{max} = V_R^{max}$.

B. The mappings minimizing $H(S)$.

The mappings where $H(S)$ is minimum are those where $V_S = 1$, i.e. only one word has non-zero degree (Ferrer-i-Cancho & Díaz-Guilera, 2007), and thus $\omega_i < 2$, which in turn implies that the principle of contrast is satisfied. To understand the relationship between the minimization $H(S)$ and the principle of contrast, notice that (a) the principle of contrast is equivalent to $H(S|R)=0$ and (b) pressure to reduce $H(S)$ implies indirect pressure for satisfying the principle of contrast because $H(S) \geq H(S|R)$ by the non-negativity of mutual information (Ferrer-i-Cancho & Díaz-Guilera, 2007). In sum, minimum $H(S)$ implies the principle of contrast but that principle is not a sufficient condition for minimum $H(S)$. Figs. 1 (a) and (b) show mappings that satisfy contrast with maximum $H(S)$ because all words are equally likely for having the same degree. Interestingly, the principle of mutual exclusivity is not needed by minimum $H(S)$.

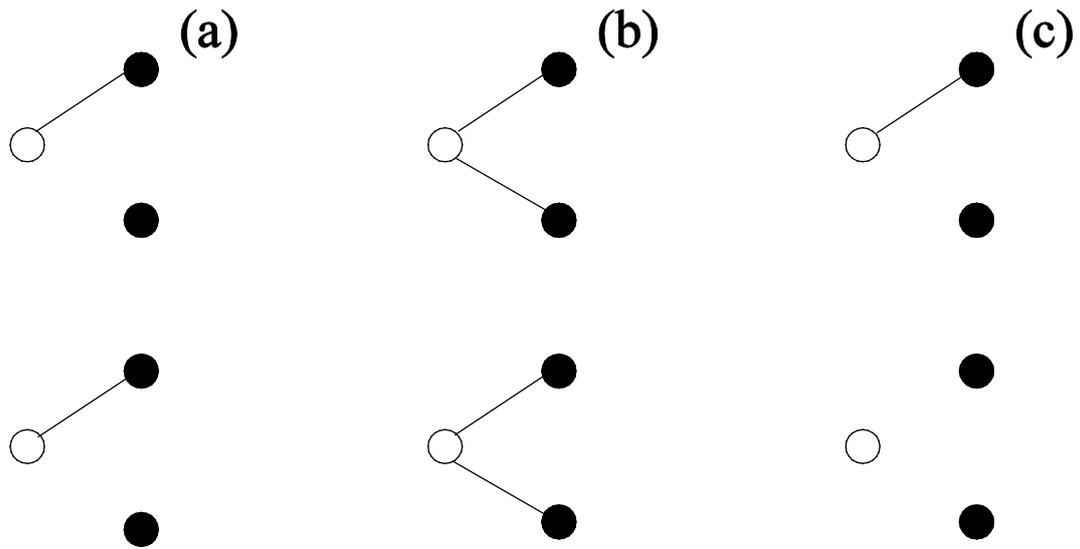

Figure 1. Toy examples of semantic memories with two words and four meanings. White and black circles stand for words and meanings, respectively. (a) A network configuration with unlinked meanings. (b) A network configuration without unlinked meanings. (c) A network configuration with just one link.